\documentclass[conference]{IEEEtran}
\IEEEoverridecommandlockouts
\usepackage{cite}
\usepackage{amsmath,amssymb,amsfonts}
\usepackage{algorithmic}
\usepackage{graphicx}
\usepackage{textcomp}
\usepackage{caption}
\usepackage{svg}
\usepackage{booktabs}
\usepackage{subcaption}
\usepackage{xcolor}
\usepackage{dsfont}

\def\BibTeX{{\rm B\kern-.05em{\sc i\kern-.025em b}\kern-.08em
    T\kern-.1667em\lower.7ex\hbox{E}\kern-.125emX}}
\begin{document}

\title{A Machine Learning Approach for Simultaneous Demapping of QAM and APSK Constellations}

\author{\IEEEauthorblockN{Arwin Gansekoele\IEEEauthorrefmark{1}\IEEEauthorrefmark{2}, Alexios Balatsoukas-Stimming\IEEEauthorrefmark{3}, Tom Brusse\IEEEauthorrefmark{4},\\ Mark Hoogendoorn\IEEEauthorrefmark{2}, Sandjai Bhulai\IEEEauthorrefmark{2} and Rob van der Mei\IEEEauthorrefmark{1}\IEEEauthorrefmark{2}}
\IEEEauthorblockA{\IEEEauthorrefmark{1}Centrum Wiskunde \& Informatica, Amsterdam, the Netherlands}
\IEEEauthorblockA{\IEEEauthorrefmark{2}Vrije Universiteit, Amsterdam, the Netherlands}
\IEEEauthorblockA{\IEEEauthorrefmark{3}Eindhoven University of Technology, Eindhoven, the Netherlands}
\IEEEauthorblockA{\IEEEauthorrefmark{4}Ministry of Defence, The Hague, the Netherlands}
\IEEEauthorblockA{Email: arwin.gansekoele@cwi.nl}
}

\maketitle

\begin{abstract}
As telecommunication systems evolve to meet increasing demands, integrating deep neural networks (DNNs) has shown promise in enhancing performance. However, the trade-off between accuracy and flexibility remains challenging when replacing traditional receivers with DNNs. This paper introduces a novel probabilistic framework that allows a single DNN demapper to demap multiple QAM and APSK constellations simultaneously. We also demonstrate that our framework allows exploiting hierarchical relationships in families of constellations. The consequence is that we need fewer neural network outputs to encode the same function without an increase in Bit Error Rate (BER). Our simulation results confirm that our approach approaches the optimal demodulation error bound under an Additive White Gaussian Noise (AWGN) channel for multiple constellations. Thereby, we address multiple important issues in making DNNs flexible enough for practical use as receivers.
\end{abstract}


\section{Introduction}
In the modern era, telecommunication systems have served as the foundation for an unprecedented revolution in information sharing, connecting people and systems across the globe in real time. From enabling seamless international business transactions to facilitating critical emergency responses, these intricate networks have become indispensable to our daily lives. However, as we continue to scale these systems to meet the ever-growing demands for higher data rates, lower latency, and more reliable connections, a myriad of challenges emerge. These range from network congestion and channel distortion to the efficient allocation of resources, and pose significant obstacles to the continued expansion and optimization of telecommunication infrastructures. To continue addressing these challenges, the use of data-driven solutions such as DNNs has emerged as a promising approach \cite{zhang2019icsta, ozpoyraz2022, you2021towards}. 

One such approach is to replace a traditional receiver with a DNN to improve the performance. There are multiple ways that differ in which components are replaced. First, one could replace the entire sender-receiver pipeline with a DNN \cite{oshea2017ae, dorner2018ijstsp, xie2021huiqiang}. Second, one could perform only channel estimation and equalization using DNNs \cite{soltani2019icl, balevi2020, wang2021yazheng}. Finally, it is possible to replace the entire receiver but leave the transmitter untouched \cite{honkala2021itwc, zhao2021zhongyuan, korpi2021, raviv2023tomer, raviv2023tomer2}. These approaches exist on a spectrum where replacing more components with a DNN leads to increased accuracy at the expense of reduced flexibility.

To better illustrate this flexibility issue, we focus on the third case, where the receiver is replaced with a DNN. Some of the most important tasks of a receiver generally include channel equalization followed by symbol demapping/demodulation. A DNN can model a receiver by taking the received sequence, possibly augmented with pilots, and directly outputting the bit log-likelihood ratios (LLRs) \cite{honkala2021itwc}. By using a learning-based approach, \cite{honkala2021itwc} were able to perform the demapping accurately with minimal numbers of pilots.

However, traditional demappers can adjust many settings as needed. For example, one could adjust the bit mapping of the symbols in a constellation or even demap a different constellation entirely. Using multiple-bit mappings may not be necessary for equally-spaced constellations such as QAM, but generalized constellations may permit multiple different bit mappings.

One would need a new neural network for every variant by hardcoding the mapping and constellation. For example, for the DVB-S2x standard \cite{morello2003}, one would need at least 118 different DNNs to model all settings. Implementing a single neural network in telecommunication systems is already complex; scaling that to 118 different networks significantly amplifies the challenges. Such an approach is also wasteful, as the problems of channel estimation and equalization are often similar across constellations. Thus, sharing the same DNN would be beneficial for these tasks. While \cite{honkala2021itwc} includes a scheme to model all variants of $4^n$-QAM with one DNN, there is, to our knowledge, no further work done for a generalizable solution to this problem.

To that end, we propose a probabilistic framework that makes it possible for a neural network demapper to operate on multiple constellations. Our framework is provably more general than previous neural demapper by including modularity without compromising performance. Our contributions are as follows. 
\begin{enumerate}
    \item We provide a generalization of \cite{honkala2021itwc} in our framework that allows neural demappers to predict bit LLRs based on a bit mapping that is not hard-coded. This allows neural demappers using our framework to be as flexible as traditional demappers.
    \item We add a representation and associated mapping to the framework. We show that by adding these, our framework becomes a generalization of methods that directly model bit LLRs with a DNN. It would thus be as accurate as these methods while being more flexible.
    \item We demonstrate how our framework allows us to use a hierarchical structure in APSK to model multiple APSK constellations for neural demappers. We do so with fewer bits without compromising accuracy.
    \item We empirically evaluate our framework and show that a neural demapper trained on multiple constellations simultaneously approaches the optimal demodulation error bound under AWGN for all constellations.
\end{enumerate}

\section{Methodology}

Symbol demapping or demodulation is the conversion of received signals back into digital data. Generally, a demapper computes an LLR or makes a hard decision by comparing the received I/Q samples to a set of constellation points. Given parameters $\theta$ and a received sequence $\hat{x}\in \mathbb{C}^L$, one could define this process using a DNN $f^{(\text{LLR})}$ as

\begin{equation}
     \left( f^{\text{(LLR)}}_{\theta}(\hat{x}) \right)_{ij}
    \label{eq:llr-receiver} = \log \frac{P(b_i = 1 \mid \hat{x}_j, \theta)}{P(b_i = 0 \mid \hat{x}_j, \theta)}.
\end{equation}

The neural network takes the sequence $\hat{x}$ and outputs the LLR for each element of the sequence. The number of bits $B$ here depends on the number of symbols in the original constellations. The network outputs the negative LLR for convenience, as a convenient property of the negative LLR is that the sigmoid function ($\sigma$) gives
\begin{equation}
    \sigma\left(\log \frac{P(b_i = 1 \mid \hat{x}_j, \theta)}{P(b_i = 0 \mid \hat{x}_j, \theta)}\right) = P(b_i = 1 \mid \hat{x}_j, \theta).
\end{equation}

This probability can be optimized using binary cross-entropy. Equation~\eqref{eq:llr-receiver} can be used jointly to demap and offset the channel effect as most of the complexity is hidden behind the learned parameters $\theta$.

This is where the problem of inflexibility we put forward earlier stems from. As the DNN is highly non-convex, it is not straightforward to adjust the network to, e.g., use a different bit mapping. To support a new bit mapping $\mathcal{M}_b$, one would need access to the underlying symbols. Overall, that is why we propose adjustments to this formulation.

\subsection{Mapping Independence}

While defining an optimal, standard bit mapping for QAM is straightforward, this is not the case for all constellations. That is why it is desirable that a demapper can perform any arbitrary mapping from symbols to bit LLRs. We model the probability of a received sample $\hat{x}$ being symbol $s \in S$ directly with a neural network to achieve this. Here, a symbol corresponds to the label of a constellation point, and $S$ is the set of all symbols in a constellation. We define the symbol DNN $f^{\text{(symbol)}}_\theta$ as

\begin{equation}  \left(f^{\text{(symbol)}}_\theta(\mathbf{\hat{x}})  \right)_{ij} = P(s_i \in S \mid \hat{x}_j, \theta).
    \label{eq:symbol-receiver}
\end{equation}

As the DNN now directly models the labels of the constellation points, we can explicitly define the bit mapping $\mathcal{M}_b$ as a function that maps symbol probabilities to bit probabilities. We define the resulting bit probabilities as

\begin{equation}
    \begin{aligned}
         P(b_i = 1 \mid \hat{x}_j, \theta) =  &\sum_{\substack{s \in S \\ \left(\mathcal{M}_b(s)\right)_i = 1}} P\left(s \mid \hat{x}_j, \theta\right).
    \end{aligned}
    \label{eq:symbol-to-llr}
\end{equation}

Equation~\eqref{eq:symbol-to-llr} uses the fact that the probability $P(b_i = 1\mid\hat{x}_j)$ corresponds to the probability that $\hat{x}_j$ should be demapped as one of the symbols that map to bit $i$. It can thus be computed by summing over all symbol probabilities where $\left(\mathcal{M}_b(s)\right)_i = 1$. Note that $\mathcal{M}_b$ is a function that maps a symbol to a bit string and thus $\left(\mathcal{M}_b(s)\right)_i$ refers to the $i$'th bit of symbol $s$ under bit mapping $\mathcal{M}_b$.

This formulation can be numerically unstable, as it often means adding many small values to a number close to $1$. Using the \emph{LogSumExp} (LSE) trick alongside clipping the probabilities so they cannot be larger than $1$ is sufficient to guarantee stability at inference time, but we have found it sub-optimal for training. To address this, note that an important feature of the LSE function is that it is a smooth approximation of the maximum function. As the training error goes to $0$, at most one term in this series does not go to $0$. Consequently, optimizing a stable pseudo-objective is one approach to deal with instability. We denote $\hat{s}$ the ground-truth symbol for $\hat{x_j}$ and $\mathds{1}$ the indicator function which is equal to $1$ when $\hat{s} = s$ and $0$ otherwise. The approximate bit probability is then
\begin{equation}
        P(b_i = 1 \mid \hat{x}_j, \theta) \approx \!\! \sum_{\substack{s \in S \\ \left(\mathcal{M}_b(s)\right)_i = 1}} \mathds{1}[s = \hat{s}] P\left(s \mid \hat{x}_j, \theta\right).
    \label{eq:pseudo-symbol-to-llr}
\end{equation}

Intuitively, \eqref{eq:pseudo-symbol-to-llr} only considers the probability of the ground-truth symbol during optimization. This addresses the instability, as we no longer need to apply a sum over probabilities. In the limit where the loss is $0$, \eqref{eq:symbol-to-llr} and \eqref{eq:pseudo-symbol-to-llr} are equivalent since the loss is minimal when the correct symbol is demapped. Note that we still use \eqref{eq:symbol-to-llr} at test time.

\subsection{${4^n}$-QAM Representation}

\begin{figure}[t]
    \centering  \includegraphics[width=0.72\linewidth]{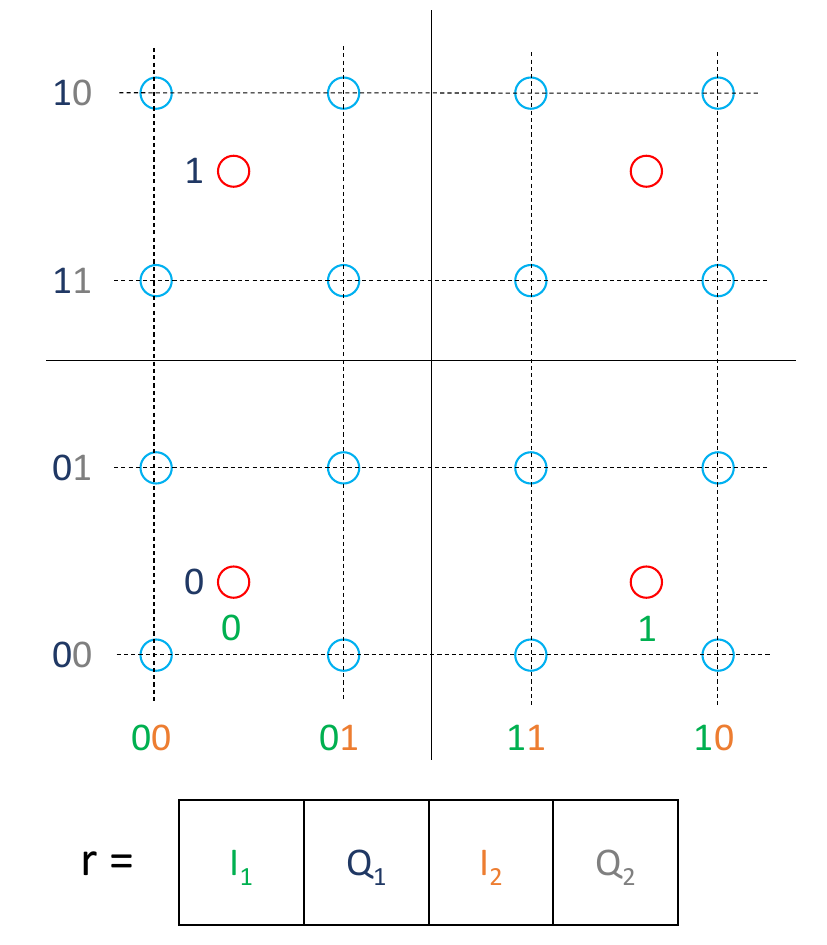}
    \caption{An example of QPSK and 16QAM overlaid. The representation is determined by counting from left to right and bottom to top in Gray code. Afterwards, interleave the in-phase and quadrature components to get a hierarchy. The representation values match the position in the representation by colour.}
    \label{fig:qam-representation}
\end{figure}

While mapping independence is an important property to have, optimization of the network becomes exponentially harder through this formulation. If the network previously had to predict $8$ bits, it now has to predict $256$ symbols. Many constellation shapes have large numbers of symmetries, so we are optimizing a model to predict with $256$ degrees of freedom where $8$ is sufficient. A constellation such as $4^n$-QAM can be represented in a highly compact manner due to its square structure. We demonstrate this in Fig.~\ref{fig:qam-representation}. Intuitively, the square structure makes it possible to encode every point in, e.g., $16$-QAM using two sets of four lines. This can thus be encoded using two sets of two bits, which we interleave to get a hierarchy. One can confirm this hierarchy with the overlaid QPSK constellation, which forms a hierarchy with $16$-QAM due to the interleaving and Gray code. 

Knowing this, we adjust the network again to ensure it takes advantage of this hierarchy while retaining the mapping independence. To do so, we formulate a hierarchical representation $r$ with associated mapping $\mathcal{M}_r$ that maps symbols to the representation. While representation $r$ can be of any type and dimension, we opted for a binary representation as it is a natural fit for $4^n$-QAM. The desired representation $r^{(4^n\text{-QAM})}$ is straightforward in this framework. As \cite{honkala2021itwc} have already observed that $4^n$-QAM corresponds to a hierarchical classifier, we can take the representation $\mathcal{M}_r$ such that it is an extension of $\mathcal{M}_b$ for all $4^n$-QAM constellations. Given this representation mapping, we propose the DNN $f_\theta^{(\text{repr})}$ that predicts the LLRs of this representation as

\begin{equation}
     \left( f_\theta^{(\text{repr})}(\mathbf{\hat{x}}) \right)_{ij} = \log \frac{P(r_i = 1 \mid \hat{x}_j, \theta)}{P(r_i = 0 \mid \hat{x}_j, \theta)}.
    \label{eq:repr-receiver}
\end{equation}

We use \eqref{eq:repr-receiver} for our simulations. Having defined both the DNN and representation mapping, We can compute the probability of a symbol $s$ by multiplying the probabilities of the hierarchical classifier based on mapping $\mathcal{M}_r$ as

\begin{equation}
    \log P(s \in S \mid \hat{x}_j) = \sum_{k = 1}^R \log P(r_i = (\mathcal{M}_r(s))_k \mid \hat{x}_j, \theta).
    \label{eq:hierarchical-prob}
\end{equation}

The advantage of this is that we can now split the different functions required of a demapper without compromising the expressiveness of our neural network. If we take $\mathcal{M}_r$ such that it is an extension of $\mathcal{M}_b$ for some constellation, it follows that $P(b_i=1\mid\hat{x}_j) = P(r_i=1\mid\hat{x}_j), \forall i\in \mathcal{D}(\mathcal{M}_b)$. Here, $\mathcal{D}(\mathcal{M}_b)$ refers to the domain of the bit mapping IE. the indices of the bit string $b$. This result implies that our formulation for $4^n$-QAM is a generalization of directly modelling the LLRs with a neural network. Our framework should thus be just as accurate as models directly optimized on the LLR while able to adjust the bit mapping. As a side note, while the remaining QAM constellations do not follow this hierarchy, we can still add them to the representation. Doing so gives us a shared representation for $4^n$-QAM and separate representations for $2*4^n$-QAM to get a representation for all QAM constellations.

\subsection{APSK Representation}

\begin{figure}[t]
    \centering
\includegraphics[width=0.72\linewidth]{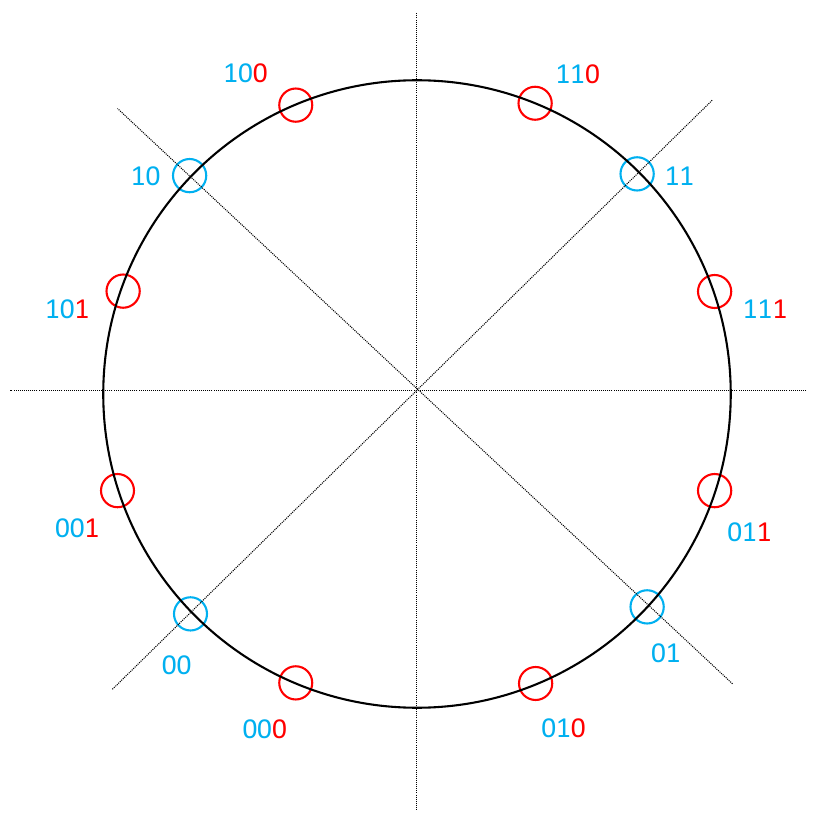}
    \caption{A QPSK (blue) and offset 8-PSK (red) constellation overlaid. The latter constellation hierarchically relates to the former, as by adding one more bit (red) to the two bits (blue), we can split the quadrants in two and move from QPSK to offset $8$-PSK.}
    \label{fig:apsk-representation}
\end{figure}

The advantage of separating $\mathcal{M}_r$ and $\mathcal{M}_b$ is not immediately apparent for $4^n$-QAM. However, the family of $4^n$-QAM is anomalous because most families of constellations do not permit such a direct equivalence between the bit mapping and a hierarchical classifier. A notable example of this is the family of circular APSK constellations. This family usually consists of a set of circles with constellation points located on these circles. These configurations can vary greatly in terms of the number of circles, spacing between circles, and points on each circle. If we take the DVB-S2x standard \cite{morello2003} of APSK constellations, for example, some common constellations are (4+12)-APSK, (8+8)-APSK, or (4+12+16)-APSK. Here, a (4+12)-APSK constellation, for example, refers to a shape with two circles where the inner and outer circles have $4$ and $12$ constellation points, respectively.

Most of these constellations have different bit mappings that are also not generally hierarchically related despite many similarities across APSK constellations that could be used to create a hierarchical classifier. All symbols are located on circles, and symbols are usually equally spaced. Furthermore, while any phase offset is possible, many circles have a phase offset of $\pi/M$ with $M$ the number of symbols on that circle. In particular, an interesting hierarchy occurs for this specific phase offset. To illustrate this, we have drawn both QPSK and 8-PSK with an offset of $\pi/8$ in Fig.~ \ref{fig:apsk-representation}. If we wanted to identify the symbols based on decision boundaries, we could use the same decision boundaries for both constellations. In both constellations, one could use two bits to identify which of the four quadrants the point lies in. For 8-PSK with offset, one only needs to add one more bit to identify which point in the quadrant it is. This hierarchy allows us to reduce the size of our APSK representation.

In fact, given this offset of $\pi/M$, this hierarchy holds for multiple series of points. If we have $4d$ constellation points on a circle with $d = 1, 3, 5, 7, \dots$ an odd number, then every circle with $2^n \times 4d$ points with $n = 1, 2, \dots$ is part of this hierarchy. For each of these series, we get the effect that doubling the number of points per quadrant comes down to splitting each decision region in two. Using this hierarchical relationship, we can model multiple APSK constellations using a smaller shared representation without losing accuracy. Note that we still need separate representations for different values of $d$, as each series forms its own hierarchy.

However, this hierarchy only captures part of the constellation. For example, the amplitudes of the circles differ for each constellation. We need separate representations for each constellation and code rate to adequately capture this. With Gray coding, we label the circles from the smallest amplitude to the largest. Adding this component, our representation is complete for constellations that satisfy the hierarchy mentioned earlier.

Not all APSK constellations have an offset of $\pi/8$. Some APSK constellations are even irregular to a degree where none of the assumptions made earlier apply. An important example is 8-PSK, which has an offset of $0$. To handle these constellations, we simply append their bits to the overall representation such that $\mathcal{M}_r$ is an extension of $\mathcal{M}_b$ for these constellations.

Finally, we mask any bits not used in the calculation. To conclude, we can now model any arbitrary APSK constellation with fewer bits than previously required through the combination of constellation-specific bits and a set of general hierarchies. It is also possible to combine the QAM and APSK representations by concatenating both. Interestingly, combining both allows for additional representation sharing. The first two bits in the $4^n$-QAM representation correspond to the quadrant as well and thus can be shared with the APSK representation.



\section{Simulation Results}

\begin{figure}[t]
    \centering
    \includegraphics[width=0.8\linewidth]{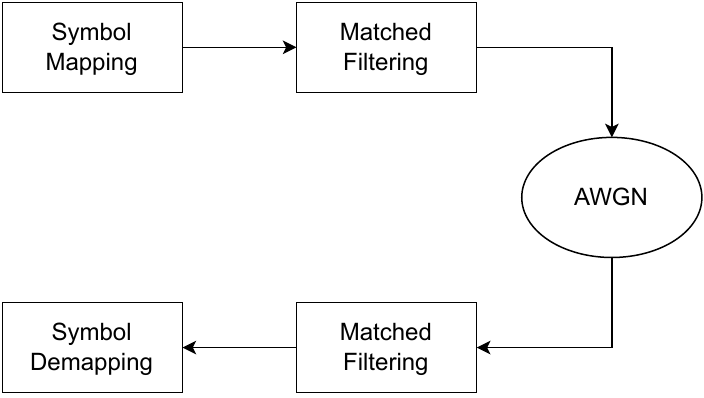}
    \caption{The system model used to evaluate our approach. For symbol demapping, we use a hard-decision demapper as a baseline and compare it to our DNN.}
    \label{fig:awgn-channel}
\end{figure}

\begin{table}[t]
\centering
\caption{Hyperparameters used for the simulations.}
\label{tab:hyperparameters}
\begin{tabular}{ll}
\toprule
\textbf{Name} & \textbf{Value} \\
\midrule
\# of Sequences &  $10000$ \\
Block Size & $2048$\\
SNR Range (dB) & $-5$ - $30$\\
Filter Rolloff & $0.25$\\
Filter Span & $10$\\
Optimiser & AdamW\cite{KingBa15, loshchilov2017decoupled} \\
Learning Rate & 0.003 \\
Batch Size & $64\times $ \#Datasets \\
Number of Runs & 3\\
Max Epochs & 100 \\
\bottomrule
\end{tabular}
\end{table}


To evaluate the performance of our method, we have opted to simulate data over a simple AWGN channel to better understand how our method behaves under various combinations of constellations. As an AWGN channel permits a simple optimal hard-decision demodulation rule, we use that as a bound to benchmark our method. The simulation pipeline is depicted in Fig.~\ref{fig:awgn-channel}.

To perform our experiments, we use a simple DNN receiver. It consists of two hidden layers of size $256$ each followed by a ReLU activation function and a batch normalization layer. It has a linear output layer that generates the representation, after which we apply our method to generate the bit LLRs. We denote the settings used to run our experiments in Table~\ref{tab:hyperparameters}. Out of the 10,000 sequences per constellation, we use 1,000 for testing, 1,000 for validation, and 8,000 for training. We found these settings sufficient to train a good neural demapper in our validation. 

For the AWGN channel, we sampled uniformly at random from the SNR range in Table~\ref{tab:hyperparameters}. Each sequence has a different SNR and thus the DNN demapper should learn how to correct for the noise level. We generate our APSK constellations based on the DVB-S2x specification \cite{morello2003}. These constellations are specified by the configuration and code rate. For example, the constellation $64$-APSK-7/9 is the APSK constellation with $64$ constellation points associated with the code rate 7/9 under this specification. The code rate generally specifies a different distance between the circles in the APSK constellation. We take a diverse subset of the $118$ constellations for our experiments.

With our simulations, we wanted to evaluate how accurate our joint method is compared to what is theoretically achievable. We split up the results into three sections.

\begin{enumerate}
    \item First, we trained a network on $256$-QAM and validated that it transfers to $64$-QAM, $16$-QAM, and QPSK.
    \item Second, we trained a network on multiple QAM and APSK constellations and evaluated its performance.
    \item Finally, we evaluated whether the model from the second experiment mimicked the decision boundaries previously defined in Fig.~\ref{fig:apsk-representation}.
\end{enumerate}

\subsection{QAM Generalisability}
\begin{figure}[t]
    \centering
    \includegraphics[width=\linewidth]{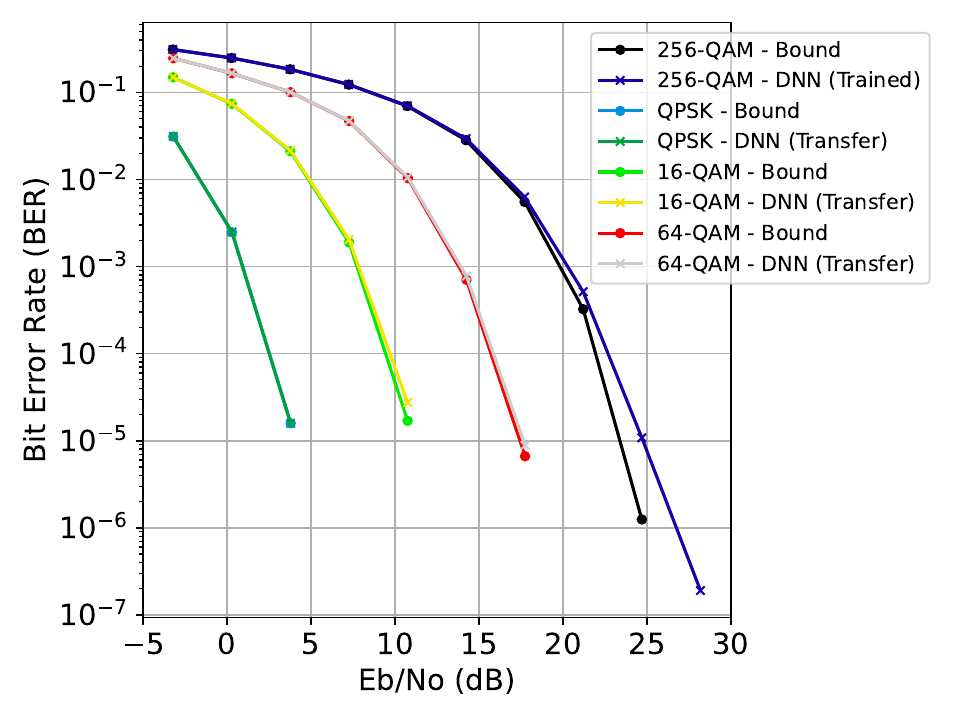}
    \caption{The BER of a DNN trained on $256$-QAM compared to the optimal hard decision bound in our setup. The (trained) and (transferred) tags imply whether the model saw the constellation during training or not.}
    \label{fig:qam-generalization}
\end{figure}

While the authors in \cite{honkala2021itwc} proposed an architecture that can support multiple QAM constellations, they did not train one model to support all simultaneously. That is why, first, we look into whether our method can generalize from one larger QAM constellation to smaller constellations. To do so, we have trained one DNN on $256$-QAM and evaluated it on QPSK, $16$-QAM, and $256$-QAM. We report the results in Fig.~\ref{fig:qam-generalization}.

Interestingly, the model approaches the boundary set by the hard-decision baseline for all constellations. However, the model was only trained on the $256$-QAM constellation. We thus confirm the original hypothesis in \cite{honkala2021itwc} and empirically show that we retain this property through our framework.

\subsection{Adding APSK Constellations}
\begin{figure}[t]
    \centering
    \includegraphics[width=\linewidth]{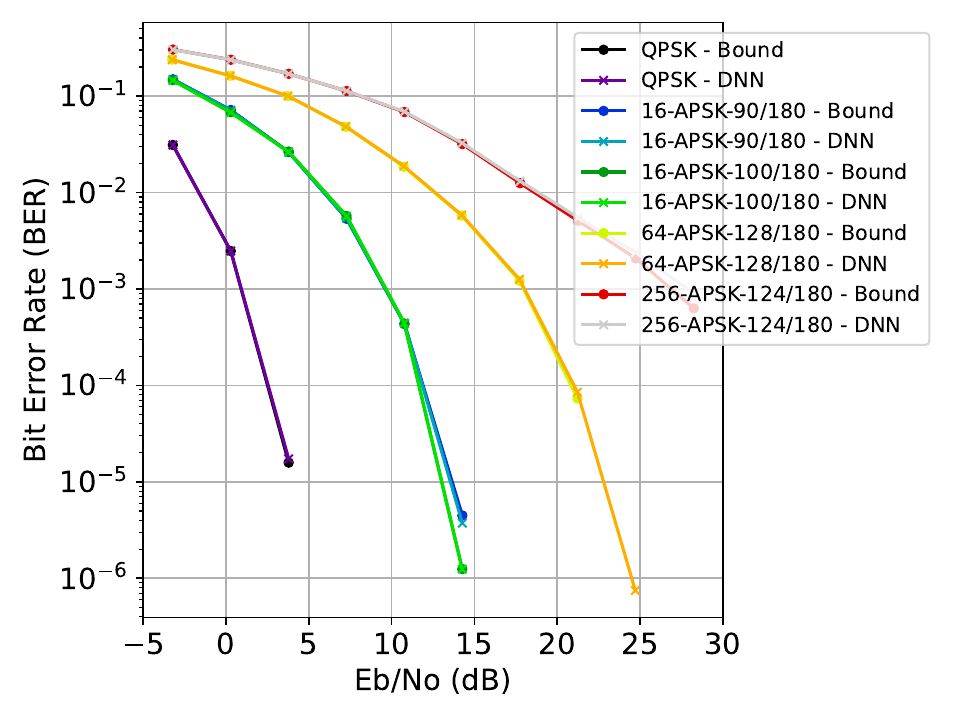}
    \caption{BER of a DNN trained on all included $4^n$-QAM and APSK constellations using our framework. Shown are the first half of the constellations trained.}
    \label{fig:transferability-1}
\end{figure}

\begin{figure}[t]
    \centering
    \includegraphics[width=\linewidth]{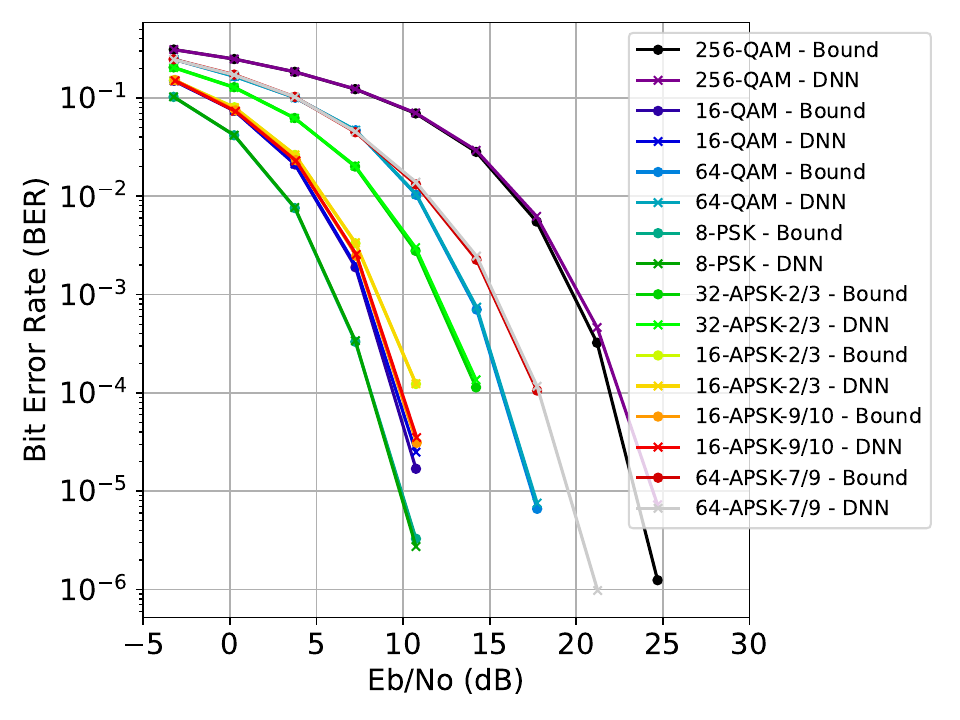}
    \caption{BER of a DNN trained on all included $4^n$-QAM and APSK constellations using our framework. Shown are the second half of the constellations trained.}
    \label{fig:transferability-2}
\end{figure}

As we have shown that our formulation allows us to train a single model for $4^n$-QAM, we now include our APSK representation for evaluation. To do so, we added nine different (A)PSK constellations from the DVB-S2x specification to the existing set of four $4^n$-QAM constellations. This set is highly diverse, giving a good overview of how our method would perform across a wide variety of constellations. We report the BER curves in Fig.~\ref{fig:transferability-1} and Fig.~\ref{fig:transferability-2}.

We find that all constellations either approach or are equal to the optimal hard-decision performance. This result is interesting, as one could think that performance would degrade when jointly optimizing over multiple objectives. However, as we only share bits between objectives that share a common goal, i.e. share a hierarchy, the multiple objectives do not actually interfere with one another. The upstream task of determining what symbol an I/Q value maps to is hypothetically also the same, and the simulation results validate this hypothesis.

Fig.~\ref{fig:transferability-1} contains one of the most interesting sets in this experiment. This is the set of QPSK, $(8+8)$-APSK-90/180, $(8+8)$-APSK-100/180, $(4 \times 16)$-APSK-128/180 and $(8 \times 32)$-APSK-124/180. Note that the adjusted notation refers to the number of circles and how many constellation points reside on each circle. All of these constellations fall under the hierarchical classifier we discussed previously, with $d=1$ as the base number (which refers to QPSK). Interestingly, all of these constellations are accurately classified with a single model despite sharing bits. Due to the sharing, we only need $12$ bits for this subset through our hierarchical representation: two shared bits for the quadrant, three bits shared by all to determine which point in the quadrant it is, and seven bits to identify which circle the constellation point lies on. If we were to model this by simply appending the required number of bits for each constellation to the representation, we would need $24$ bits. We can conclude from this set of constellations that the hierarchy we established allows sharing parameters without harming performance.

In Fig.~\ref{fig:transferability-2}, we depict the other constellations. This set consists of more distinct constellations. Notable examples are $8$-PSK, $(4+12)$-APSK-2/3 and $(8+16+20+20)$-APSK-7/9. Note that the last constellation also follows the $d=1$ hierarchy in the inner two circles. In particular, $8$-PSK does not share a single bit with any of the other constellations. Despite this, it clearly does not suffer in terms of performance. From this, we can conclude that adding separate bits for constellations that do not fit in the hierarchy can be a good solution when no hierarchy is easily identifiable. 

On a similar note, the QAM constellations also do not suffer from being trained jointly with the APSK representations. In fact, the $256$-QAM BER curve even seems slightly closer to the optimum than what was reported in Fig.~\ref{fig:qam-generalization}. It is important to note that we included the QPSK, $16$-QAM and $64$-QAM data for this training run, which we did not for Fig.~\ref{fig:qam-generalization}. This thus suggests that data from constellations that share bits can improve the performance of those constellations.

\subsection{QPSK vs offset $8$-PSK, Distribution Modelling}

\begin{figure}[t]
     \centering
     \begin{subfigure}[b]{0.24\textwidth}
         \centering
         \includegraphics[width=\linewidth]{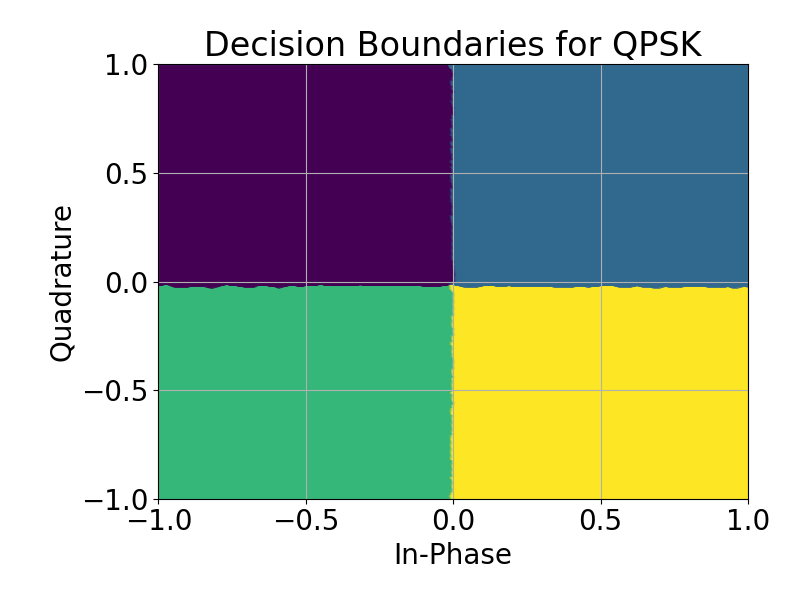}
         \caption{QPSK}
     \end{subfigure}
     \hfill
     \begin{subfigure}[b]{0.24\textwidth}
         \centering
         \includegraphics[width=\linewidth]{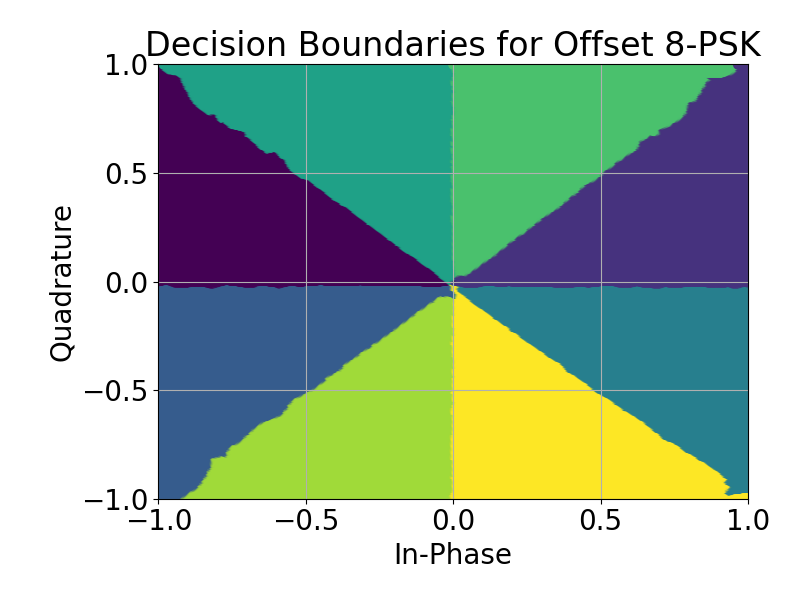}
         \caption{Offset $8$-PSK}
     \end{subfigure}
     
     \caption{The decision boundaries of the joint model for both QPSK and offset $8$-PSK. Our DNN seems to learn the optimal decision boundaries as depicted in Figure~\ref{fig:apsk-representation}.}
     \label{fig:8psk-qpsk-distribution}
\end{figure}

To get more insight into why our model performs well, we plotted the decision boundaries our DNN receiver generates in Fig.~\ref{fig:8psk-qpsk-distribution}. Here, we depict the boundaries for the symbols when considering QPSK and offset $8$-PSK as in Fig.~\ref{fig:apsk-representation}. These boundaries are based on the two quadrant representation bits and the first bit of the $d=1$ shared series. Interestingly, the decision boundaries generated by our DNN correspond roughly with the optimal decision boundaries we have previously established. Clearly, our shared representation works as the DNN models the optimal decision boundaries.

\section{Conclusion}
We proposed a framework that allows us to jointly train on and generalize to multiple types of constellations. In this framework, we included mapping independence and the capability to use hierarchical relationships in families of constellations. We found that it is not only possible to train a model jointly on multiple constellations, but that such a model approaches the hard-decision BER bound under AWGN. The framework we introduced thus opens up the possibility to efficiently model families of constellations in the context of DNNs. Our framework is modular and applicable to any deep-learning receiver pipeline. It is thus interesting to test in different settings, for more types of constellations, and under more complicated channel effects. Furthermore, application to real data and analysis of the resulting computational complexity would be interesting.

\bibliographystyle{IEEEtran}
\bibliography{IEEEexample}

\end{document}